\documentclass[preprint,12pt,authoryear]{elsarticle}
\usepackage[]{graphicx,float,latexsym,times}
\usepackage{amsfonts,amstext,amsmath,amssymb,amsthm}
\usepackage{caption2}
\usepackage{color}

\newtheorem{thm}{Theorem}
\newtheorem{cor}{Corollary}
\newtheorem{lem}{Lemma}
\newtheorem{rem}{Remark}

\newcommand{\bB}{\mbox{\boldmath {$B$}}}

\newcommand{\bI}{\mbox{\boldmath {$I$}}}

\newcommand{\bO}{\mbox{\boldmath {$O$}}}

\newcommand{\bS}{\mbox{\boldmath {$S$}}}

\newcommand{\bw}{\mbox{\boldmath {$w$}}}

\newcommand{\bx}{\mbox{\boldmath {$x$}}}

\newcommand{\bze}{\mbox{\boldmath {$0$}}}

\newcommand{\bmu}{\mbox{\boldmath $ \mu $}}
\newcommand{\bSig}{\mbox{\boldmath $ \Sigma $}}

\newcommand{\bSigma}{\mbox{\boldmath $ \Sigma $}}

\newcommand{\balpha}{\mbox{\boldmath $ \alpha $}}
\newcommand{\bsal}{\mbox{\scriptsize $\balpha$}}

\newcommand{\tr}{\mbox{\rm tr}}
\newcommand{\Var}{\mbox{\rm Var}}
\newcommand{\argmin}{\mathop{\rm argmin}\limits}
\newcommand{\argmax}{\mathop{\rm argmax}\limits}

\newcommand\CG[1]{\textcolor{black}{#1}}

\begin{document}

\begin{frontmatter}

\title{Support vector machine and its bias correction in high-dimension, low-sample-size settings}


\author[label1]{Yugo Nakayama}
\address[label1]{Graduate School of Pure and Applied Sciences, University of Tsukuba, Ibaraki, Japan}

\author[label2]{Kazuyoshi Yata}
\address[label2]{Institute of Mathematics, University of Tsukuba, Ibaraki, Japan}

\author[label2]{Makoto Aoshima\fnref{label3}}

\fntext[label3]{Institute of Mathematics, University of Tsukuba, Ibaraki 305-8571, Japan; Fax: +81-29-853-6501}
\ead{aoshima@math.tsukuba.ac.jp}

\begin{abstract}
In this paper, we consider asymptotic properties of the support vector machine (SVM) in high-dimension, low-sample-size (HDLSS) settings. 
We show that the hard-margin linear SVM holds a consistency property in which misclassification rates tend to zero as the dimension goes to infinity under certain severe conditions. 
We show that the SVM is very biased in HDLSS settings and its performance is affected by the bias directly. 
In order to overcome such difficulties, we propose a bias-corrected SVM (BC-SVM). 
We show that the BC-SVM gives preferable performances in HDLSS settings. 
We also discuss the SVMs in multiclass HDLSS settings.
Finally, we check the performance of the classifiers in actual data analyses.
\end{abstract}


\end{frontmatter}

\section{Introduction}
High-dimension, low-sample-size (HDLSS) data situations occur in many areas of modern science such as genetic microarrays, medical imaging, text recognition, finance, chemometrics, and so on. 
Suppose we have independent and $d$-variate two populations, $\pi_i,\ i=1,2$, having an unknown mean vector $\bmu_i$ and unknown covariance matrix $\bSigma_i\ (\ge \bO)$. 
We assume that $\tr(\bSig_i)/d \in (0,\infty)$ as $d\to \infty$ for $i=1,2$. 
Here, for a function, $f(\cdot)$, ``$f(d) \in (0, \infty)$ as $d\to \infty$" implies $\liminf_{d\to \infty}f(d)>0$ and $\limsup_{d\to \infty}f(d)<\infty$. 
Let $\Delta=\|\bmu_1-\bmu_2\|^2$, where $\|\cdot \|$ denotes the Euclidean norm. 
We assume that $\limsup_{d\to \infty}\Delta/d<\infty $. 
We have independent and identically distributed (i.i.d.) observations, $\bx_{i1},...,\bx_{in_i}$, from each $\pi_i$.
We assume $n_i\ge 2,\ i=1,2$. 
Let $\bx_0$ be an observation vector of an individual belonging to one of the two populations.
\CG{
We assume $\bx_0$ and $\bx_{ij}$s are independent.}
Let $N=n_1+n_2$. 

In the HDLSS context, \cite{H05}, \cite{M07} and \cite{Q10} considered distance weighted classifiers.
\cite{H08}, \cite{CH09} and \cite{AY14} considered distance-based classifiers.
In particular, \cite{AY14} gave the misclassification rate adjusted classifier for multiclass, high-dimensional data in which misclassification rates are no more than specified thresholds. 
On the other hand, \cite{AY11,AY15a} considered geometric classifiers based on a geometric representation of HDLSS data. 
\cite{AM10} considered a classifier based on the maximal data piling direction. 
\cite{AY15b} considered quadratic classifiers in general and discussed asymptotic properties and optimality of the classifies under high-dimension, non-sparse settings. 
In particular, \cite{AY15b} showed that the misclassification rates tend to $0$ as $d$ increases, i.e., 
\begin{equation}
e(i)\to 0 \ \ \mbox{as $d\to \infty$ for $i=1,2$} 
\label{1.1}
\end{equation}
under the non-sparsity such as $\Delta \to \infty$ as $d\to \infty$, where $e(i)$ denotes the error rate of misclassifying an individual from $\pi_i$ into the other class. 
We call (\ref{1.1}) ``the consistency property".
We note that a linear classifier can give such a preferable performance under the non-sparsity. 
Also, such non-sparse situations often appear in real high-dimensional data. 
See \cite{AY15b} for the details. 
Hence, in this paper, we focus on linear classifiers.

In the field of machine learning, there are many studies about the classification in the context of supervised learning. 
A typical method is the support vector machine (SVM). 
The SVM has versatility and effectiveness both for low-dimensional and high-dimensional data. 
See \cite{V00}, \cite{SS02}, \cite{H05}, \cite{H09} and \cite{QZ15} for the details. 
Even though the SVM is quite popular, its asymptotic properties seem to have not been studied sufficiently.
In this paper, we investigate asymptotic properties of the SVM for HDLSS data. 

\begin{figure}
\includegraphics[scale=0.51]{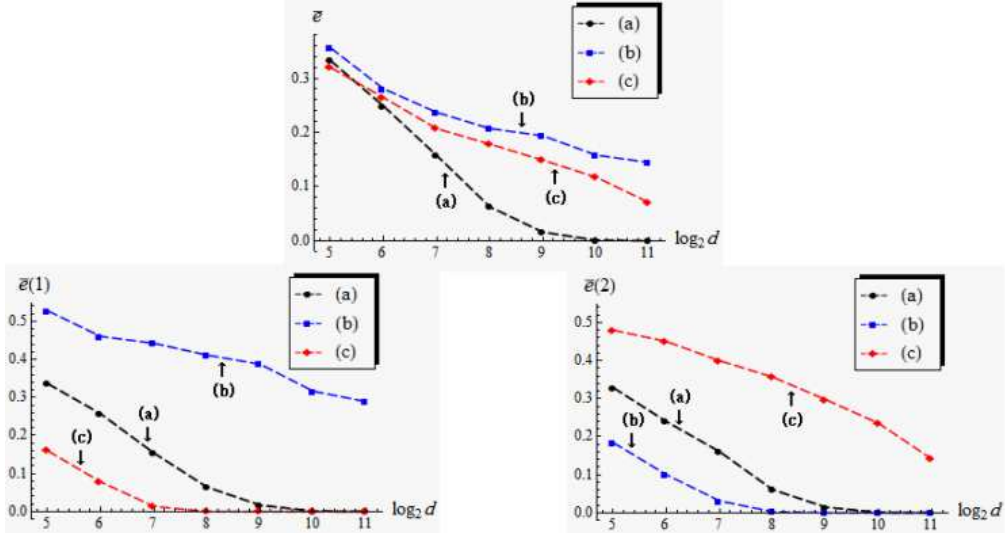} 
\caption{The performance of the SVM given by (\ref{2.4}) in HDLSS settings.
The left panel displays $\overline{e}(1)$, the right panel displays $\overline{e}(2)$ and the top panel displays $\overline{e}$. 
\CG{Their standard deviations are less than $0.0112$.}
}
\label{F1}
\end{figure}

Now, let us use the following toy examples to see the performance of the hard-margin linear SVM given by (\ref{2.4}). 
We set $N=20$ and $d=2^{s},\ s=5,...,11$. 
Independent pseudo random observations were generated from $\pi_i: N_d(\bmu_i, \bSigma_i)$, $i=1,2$.
We set $\bmu_1=\bze$ and $\bmu_2=(1/3,...,1/3)^T$, so that $\Delta=d/9$. 
We considered three cases: 
\begin{flushleft}
(a) $(n_1,n_2)=(10,10)$ and $\bSig_1=\bSig_2=\bI_d$; \\
(b) $(n_1,n_2)=(6,14)$ and $\bSig_1=\bSig_2=\bI_d$; and \\
(c) $(n_1,n_2)=(10,10)$, $\bSig_1=0.6\bI_d$ and $\bSig_2=1.4\bI_d$,
\end{flushleft}
where $\bI_{d}$ denotes the $d$-dimensional identity matrix.
Note that $\Delta>|\tr(\bSig_1)/n_1-\tr(\bSig_2)/n_2|$ for (a) to (c).
Then, from Theorem 1 in \cite{H05}, the classifier should hold (\ref{1.1}) for (a) to (c). 
We repeated 2000 times to confirm if the classifier does (or does not) classify $\bx_0\in\pi_i$ correctly and defined $P_{ir}=0\ (\mbox{or}\ 1)$ accordingly for each $\pi_i\ (i=1,2)$.
We calculated the error rates, $\overline{e}(i)= \sum_{r=1}^{2000}P_{ir}/2000$, $i=1,2$. 
Also, we calculated the average error rate, $\overline{e}=\{\overline{e}(1)+\overline{e}(2)\}/2$. 
\CG{
Their standard deviations are less than $0.0112$ from the fact that $\Var\{\overline{e}(i)\}=e(i)\{1-e(i)\}/2000\le 1/8000$.}
In Figure \ref{F1}, we plotted $\overline{e}(1)$, $\overline{e}(2)$ and $\overline{e}$ for (a) to (c).
We observe that the SVM gives a good performance as $d$ increases for (a). 
Contrary to expectations, it leads undesirable performances both for (b) and (c). 
The error rates becomes small as $d$ increases, however, $\overline{e}(1)$ and $\overline{e}(2)$ are quite unbalanced. 
We discuss some theoretical reasons in Section 2.2. 

In this paper, we investigate the SVM in the HDLSS context.
In Section 2, we show that the SVM holds (\ref{1.1}) under certain severe conditions. 
We show that the SVM is very biased in HDLSS settings and its performance is affected by the bias directly. 
In order to overcome such difficulties, we propose a bias-corrected SVM (BC-SVM) in Section 3.
We show that the BC-SVM improves the SVM even when $n_i$s or $\bSig_i$s are unbalanced as in (b) or (c) in Figure 1. 
In Section 4, we check the performance of the BC-SVM by numerical simulations and use the BC-SVM in actual data analyses.
In Section 5, we discuss multiclass SVMs in HDLSS settings. 
\section{SVM in HDLSS Settings}
In this section, we give asymptotic properties of the SVM in HDLSS settings.
Since HDLSS data are linearly separable by a hyperplane, we consider the hard-margin linear SVM. 
\subsection{Hard-margin linear SVM}
We consider the following linear classifier:
\begin{equation}
y(\bx)=\bw^T\bx+b, 
\label{2.1}
\end{equation}
where $\bw$ is a weight vector and $b$ is an intercept term. 
Let us write that $(\bx_1,...,\bx_N)=(\bx_{11},...,\bx_{1n_1},\bx_{21},...,\bx_{2n_2})$.
Let $t_j=-1$ for $j=1,...,n_1$ and $t_j=1$ for $j=n_1+1,...,N$. 
The hard-margin SVM is defined by maximizing the smallest distance of all observations to the separating hyperplane.
The optimization problem of the SVM can be written as follows:
$$
\argmin_{{\scriptsize \bw},b}\frac{1}{2}\| \bw \|^2\quad \mbox{subject to \ $t_j(\bw^T\bx_j+b)\ge 1$, $j=1,...,N$.}
$$
A Lagrangian formulation is given by 
$$
L(\bw,b;\balpha)=\frac{1}{2}|| \bw||^2-\sum_{j=1}^N\alpha_j\{t_j(\bw^T\bx_j+b)-1\}, 
$$
where $\balpha=(\alpha_1,...,\alpha_N)^T$ and $\alpha_j$s are Lagrange multipliers. 
By differentiating the Lagrangian formulation with respect to $\bw$ and $b$, we obtain the following conditions: 
$$
\bw=\sum_{j=1}^N\alpha_jt_j \bx_j \ \mbox{ and } \  
\sum_{j=1}^N\alpha_jt_j=0.
$$
After substituting them into $L(\bw,b;\balpha)$, we obtain the dual form: 
\begin{equation}
L(\balpha)=\sum_{j=1}^N\alpha_j-\frac{1}{2}\sum_{j=1}^N\sum_{k=1}^N\alpha_j\alpha_kt_jt_k \bx_j^T\bx_k.
\label{2.2}
\end{equation}
The optimization problem can be transformed into the following:
$$
\argmax_{ \bsal } L(\balpha)
$$
subject to 
\begin{equation}
\alpha_j\ge 0,\ j=1,...,N, \ \mbox{ and } \  
\sum_{j=1}^N\alpha_jt_j=0. 
\label{2.3}
\end{equation}
Let us write that 
$$
\hat{\balpha}=(\hat{\alpha}_1,...,\hat{\alpha}_N)^T=\argmax_{ \bsal }L(\balpha)\ \mbox{ subject to (\ref{2.3})}.
$$
There exist some $\bx_j$s satisfying that $t_jy(\bx_j)=1$ (i.e., $\hat{\alpha}_j\neq 0$).
Such $\bx_j$s are called the support vector.
Let $\hat{S}=\{j|\hat{\alpha}_j\neq 0,\ j=1,...,N\}$ and $N_{\hat{S}}=\# \hat{S}$, where $\# A$ denotes the number of elements in a set $A$.
The intercept term is given by
$$
\hat{b}=\frac{1}{N_{\hat{S}}}\sum_{j\in \hat{S}}\Big(t_j-\sum_{k\in \hat{S}}\hat{\alpha}_kt_k\bx_j^T\bx_k\Big). 
$$
Then, the linear classifier in (\ref{2.1}) is defined by
\begin{equation}
\hat{y}(\bx)=\sum_{k\in \hat{S}} \hat{\alpha}_k t_k \bx_k^T\bx+\hat{b}.
\label{2.4}
\end{equation}
Finally, in the SVM, one classifies $\bx_0$ into $\pi_1$ if $\hat{y}(\bx_0 )<0$ and into $ \pi_2$ otherwise. 
See \cite{V00} for the details.
\subsection{Asymptotic properties of the SVM in the HDLSS context}
\CG{In this section, we consider the case when $d\to \infty$ while $N$ is fixed.} 
We assume the following assumptions: 
\begin{description}
  \item[(A-i)]\quad $\displaystyle \frac{ \Var(\|\bx_{ik}-\bmu_i\|^2)}{\Delta^2}\to 0$ as $d\to \infty$ for $i=1,2$; 
  \item[(A-ii)]\quad $\displaystyle \frac{ \tr(\bSigma_i^2)}{\Delta^2}\to 0$ as $d\to \infty$ for $i=1,2$.
\end{description}
Note that $\Var(\|\bx_{ik}-\bmu_i\|^2)=2\tr(\bSig_i^2)$ when $\pi_i$ is Gaussian, so that (A-i) and (A-ii) are equivalent when $\pi_i$s are Gaussian.
\begin{lem}
Under (\ref{2.3}), it holds that as $d\to \infty$
$$
L(\balpha)=\sum_{j=1}^N\alpha_j- \frac{\Delta}{8}\Big(\sum_{j=1}^N\alpha_j\Big)^2\{1+o_p(1)\}
-\frac{1}{2}\Big(\tr(\bSig_1)\sum_{j=1}^{n_1}\alpha_j^2+
\tr(\bSig_2)\sum_{j=n_1+1}^{N}\alpha_j^2\Big).
$$
\end{lem}
Let $\delta=\tr(\bSig_1)/n_1+\tr(\bSig_2)/n_2$ and $\Delta_*=\Delta+\delta$. 
Under the constraint that $\sum_{j=1}^N\alpha_j=C$ for a given positive constant $C$, we can claim that  
\begin{equation}
\max_{ \bsal} \Big\{-\frac{1}{2}
\Big( \tr(\bSig_1)\sum_{j=1}^{n_1}\alpha_j^2+\tr(\bSig_2)\sum_{j=n_1+1}^{N}\alpha_j^2\Big)\Big\}
=-\frac{C^2}{8}\delta 
\label{2.5}
\end{equation}
when $\alpha_1=\cdots =\alpha_{n_1}=C/(2n_1)$ and $\alpha_{n_1+1}=\cdots =\alpha_{N}=C/(2n_2)$ under (\ref{2.3}). 
Then, \CG{by noting that $\liminf_{d\to \infty} \{\tr(\bSig_i)/(\Delta n_i)\} >0$ for $i=1,2$,} 
from Lemma 1 it holds that 
\begin{equation}
\max_{ \bsal } L(\balpha)= -\frac{\Delta_*}{8}\Big(C-\frac{4+o_p(1)}{\Delta_*} \Big)^2\{1+o_p(1)\}+\frac{2+o_p(1)}{\Delta_*}
\label{2.6}
\end{equation}
for given $C(>0)$.
Hence, by choosing $C\approx 4/\Delta_*$, we have the maximum of $L(\balpha)$ asymptotically. 
\begin{lem}
It holds that as $d\to \infty$
\begin{align*}
&\hat{\alpha}_j=\frac{2}{\Delta_*n_1}\{1+o_p(1)\}\quad \mbox{for $j=1,...,n_1$};\quad \mbox{and} \\
&\hat{\alpha}_j=\frac{2}{\Delta_*n_2}\{1+o_p(1)\}\quad \mbox{for $j=n_1+1,...,N$}.
\end{align*} 
Furthermore, it holds that as $d\to \infty$
\begin{align*}
&\hat{y}(\bx_0)=\frac{(-1)^i\Delta}{\Delta_*}+\frac{\tr(\bSig_1)/n_1-\tr(\bSig_2)/n_2}{\Delta_*}+o_p\Big(\frac{\Delta}{\Delta_*}\Big) \\
&\mbox{when $\bx_0\in \pi_i$, $i=1,2$.}
\end{align*} 
\end{lem}
\begin{rem}
From Lemma 2, all the data points are the support vectors under (A-i) and (A-ii) in the HDLSS context. 
\cite{AM10} called this phenomenon the ``data piling". 
See Sections 1 and 2 in \cite{AM10} for the details. 
\end{rem}
Let $\kappa =\tr(\bSig_1)/n_1-\tr(\bSig_2)/n_2$. 
From Lemma 2, it holds that as $d\to \infty$
\begin{equation}
\frac{\Delta_*}{\Delta}\hat{y}(\bx_0)=(-1)^i+\frac{\kappa }{\Delta}+o_p(1)
\label{2.7}
\end{equation}
when $\bx_0\in \pi_i$, $i=1,2$. 
Hence, ``$\kappa /\Delta$" is the bias term of the (normalized) SVM. 
We consider the following assumption: 
\begin{description}
  \item[(A-iii)]\quad $\displaystyle  \limsup_{d\to \infty} \frac{|\kappa|}{\Delta}<1$.
\end{description}
\begin{thm}
Under (A-i) to (A-iii), the SVM holds (\ref{1.1}). 
\end{thm}
\begin{cor}
Under (A-i) and (A-ii), the SVM holds the following properties:
\begin{align*}
&e(1)\to 1 \ \mbox{ and } \ e(2)\to 0 \ \mbox{ as $d\to \infty$ \ if \ }\liminf_{d\to \infty}\frac{\kappa }{\Delta}>1;
\quad \mbox{and} \\
&e(1)\to 0 \ \mbox{ and } \ e(2)\to 1 \ \mbox{ as $d\to \infty$ \ if \ }\limsup_{d\to \infty}\frac{\kappa }{\Delta}<-1.
\end{align*}
\end{cor}
\begin{rem}
For the SVM, \cite{H05} and \cite{QZ15} also showed (\ref{1.1}) and the results in Corollary 1 under different conditions. 
We emphasize that (A-i), (A-ii) and (A-iii) are milder than their conditions. 
Moreover, we can evaluate the bias of the SVM by using (\ref{2.7}). 
\end{rem}
We expect from (\ref{2.7}) that, for sufficiently large $d$, $e(1)$ and $e(2)$ for the SVM become small and $e(1)$ (or $e(2)$) is larger than $e(2)$ (or $e(1)$) if $\kappa/\Delta>0$ (or $\kappa/\Delta<0$). 
Actually, in Figure 1, we observe that $\overline{e}(1)$ is larger than $\overline{e}(2)$ for (b) in which $\kappa/\Delta=6/7$ and $\overline{e}(2)$ is larger than $\overline{e}(1)$ for (c) in which $\kappa/\Delta=-18/25$. 
As for (a) in which $\kappa =0$, the SVM gives a preferable performance. 
\subsection{Asymptotic properties of the SVM when both $d$ and $N$ tend to infinity}
\CG{
In this section, we give asymptotic properties of the SVM when both $d,N \to \infty$ while $N/d\to 0$.
One may consider $N=O(\log d)$ for example. 
We assume the following assumptions: 
\begin{description}
  \item[(A-i')]\quad $\displaystyle \frac{ N \Var(\|\bx_{ik}-\bmu_i\|^2)}{\Delta^2}\to 0$ as $d,N\to \infty$ for $i=1,2$; 
  \item[(A-ii')]\quad $\displaystyle \frac{N^2 \tr(\bSigma_i^2)}{\Delta^2}\to 0$ as $d,N\to \infty$ for $i=1,2$;
   \item[(A-iv)]\quad $\displaystyle \liminf_{d,N\to \infty} \frac{\tr(\bSigma_i)}{ \Delta n_i}> 0$ for $i=1,2$.
\end{description}
Note that $\Delta^2/\tr(\bSigma_i^2)=O(d)$ from the facts that $\limsup_{d\to \infty}\Delta/d<\infty$ 
and $\tr(\bSig_i)/d\in (0,\infty)$ as $d\to \infty$ for $i=1,2$. 
Thus, $N=o(d^{1/2})$ when (A-ii') is met. 
\begin{lem}
Under (A-i'), (A-ii') and (A-iv), it holds that as $d,N\to \infty$
$$
\hat{y}(\bx_0)=\frac{(-1)^i\Delta}{\Delta_*}+\frac{\kappa}{\Delta_*}+o_p\Big(\frac{\Delta}{\Delta_*}\Big) \ \mbox{ when $\bx_0\in \pi_i$ for $i=1,2$.}
$$
\end{lem}
\begin{cor}
Under (A-i'), (A-ii') and (A-iv), the SVM holds the following properties:
\begin{align*} 
&e(1)\to 0 \ \mbox{ and } \ e(2)\to 0 \ \mbox{ as $d,N\to \infty$ \ if \ } \limsup_{d,N\to \infty} \frac{|\kappa|}{\Delta}<1;
  \\
&e(1)\to 1 \ \mbox{ and } \ e(2)\to 0 \ \mbox{ as $d,N\to \infty$ \ if \ }\liminf_{d,N\to \infty}\frac{\kappa }{\Delta}>1;
\quad \mbox{and} \\
&e(1)\to 0 \ \mbox{ and } \ e(2)\to 1 \ \mbox{ as $d,N\to \infty$ \ if \ }\limsup_{d,N\to \infty}\frac{\kappa }{\Delta}<-1.
\end{align*}
\end{cor}
}
\section{Bias-Corrected SVM}
As discussed in Section 2.2, if $\liminf_{d\to \infty}|\kappa|/\Delta>0$, the SVM gives an undesirable performance. 
From Corollary 1, if $\liminf_{d\to \infty}|\kappa|/\Delta>1$, one should not use the SVM. 
In order to overcome such difficulties, we consider a bias correction of the SVM.  

We estimate $\bmu_i$ and $\bSigma_i$ by $\overline{\bx}_{in_i}=\sum_{j=1}^{n_i}{\bx_{ij}}/{n_i}$ and $\bS_{in_i}=\sum_{j=1}^{n_i}(\bx_{ij}-\overline{\bx}_{in_i})(\bx_{ij}-\overline{\bx}_{in_i})^T/(n_i-1)$.  
We estimate $\Delta_*$ by $\hat{\Delta}_*=\|\overline{\bx}_{1n_1}-\overline{\bx}_{2n_2} \|^2$.
Note that $E(\hat{\Delta}_*)=\Delta_*$. 
Let $\hat{\kappa}=\tr(\bS_{1n_1})/n_1-\tr(\bS_{2n_2})/n_2 $. 
Note that $E(\hat{\kappa})=\kappa$. 
\CG{First, we consider the case when $d\to \infty$ while $N$ is fixed.} 
\begin{lem}
Under (A-i) and (A-ii), it holds that as $d\to \infty$ 
$$
\frac{\hat{\kappa}}{\hat{\Delta}_*}=
\frac{{\kappa}}{{\Delta}_*}+o_p\Big(\frac{\Delta}{\Delta_*}\Big).
$$
\end{lem}
Now, we define the bias-corrected SVM (BC-SVM) by
\begin{equation}
\hat{y}_{BC}(\bx_0)=\hat{y}(\bx_0)-\frac{\hat{\kappa}}{\hat{\Delta}_*}, 
\label{3.1}
\end{equation}
where $\hat{y}(\bx_0)$ is given by (\ref{2.4}).
In the BC-SVM, one classifies $\bx_0$ into $\pi_1$ if $\hat{y}_{BC}(\bx_0)<0$ and into $\pi_2$ otherwise. 

By combining (\ref{2.7}) with Lemma 4, under (A-i) and (A-ii), it holds that as $d\to \infty$
\begin{equation}
\frac{\Delta_*}{\Delta}\hat{y}_{BC}(\bx_0)=(-1)^i+o_p(1)
\label{3.2}
\end{equation}
when $\bx_0\in \pi_i$, $i=1,2$. 
\begin{thm}
Under (A-i) and (A-ii), the BC-SVM holds (\ref{1.1}). 
\end{thm}
\begin{rem}
One should note that the BC-SVM has the consistency property without (A-iii). 
\cite{CH09} considered a different bias correction for the SVM. 
They showed the consistency property under some stricter conditions than (A-i) and (A-ii). 
\end{rem}
\begin{rem}
\cite{AY14} considered the distance-based classifier as follows: 
One classifies an individual into $\pi_1$ if $y_{AY}(\bx_0)<0$ and into $\pi_2$ otherwise, 
where $y_{AY}(\bx_0)=\{\bx_0-(\overline{\bx}_{1n_1}+\overline{\bx}_{2n_2})/2\}^T(\overline{\bx}_{2n_2}-\overline{\bx}_{1n_1})-\tr(\bS_{1n_1})/(2n_1)+\tr(\bS_{2n_2})/(2n_2)$. 
Then, from Theorem 1 in \cite{AY14}, 
under (A-ii), it holds that as $d\to \infty$
$$
(2/\Delta)y_{AY}(\bx_0)=(-1)^i+o_p(1)
$$
when $\bx_0\in \pi_i$, $i=1,2$. 
\end{rem}
\CG{
When both $d,N \to \infty$, we have the following result. 
\begin{cor}
Under (A-i'), (A-ii') and (A-iv), it holds for the BC-SVM that $e(i)\to 0$ as $d,N\to \infty$ for $i=1,2$. 
\end{cor}
}
\section{Performances of Bias-Corrected SVM}
In this section, we check the performance of the BC-SVM both in numerical simulations and actual data analyses. 
\subsection{Simulations}

First, we checked the performance of the BC-SVM by using the toy examples in Figure 1. 
Similar to Section 1, we calculated the error rates, $\overline{e}(1)$, $\overline{e}(2)$ and $\overline{e}$, by 2000 replications and plotted the results in Figure 2. 
We laid $\overline{e}(1)$, $\overline{e}(2)$ and $\overline{e}$ for the SVM by borrowing from Figure 1. 
As expected theoretically, we observe that the BC-SVM gives preferable performances even for (b) and (c) in which $\liminf_{d\to \infty}|\kappa|/\Delta>0$. 
\begin{figure}
\includegraphics[scale=0.4]{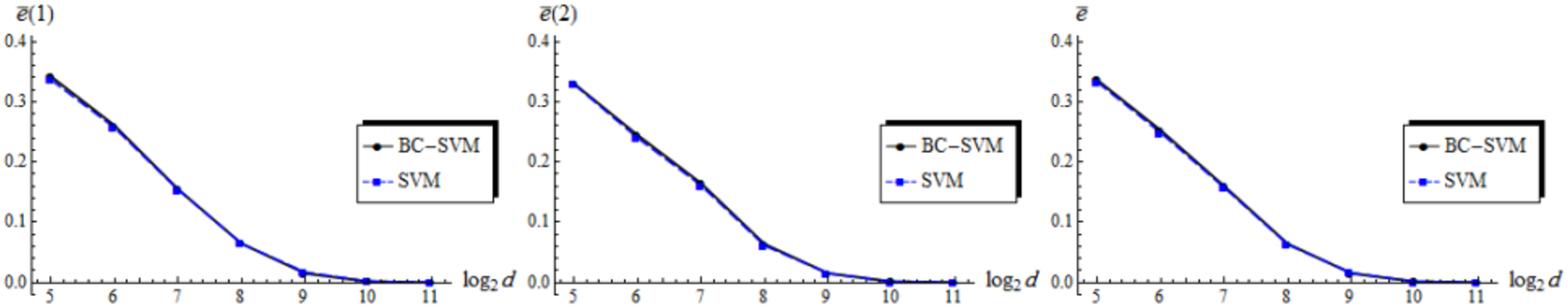} \\[-1mm]
(a) $(n_1,n_2)=(10,10)$ and $\bSig_1=\bSig_2=\bI_d$ (i.e., $\kappa=0$) \\[2mm]
\includegraphics[scale=0.4]{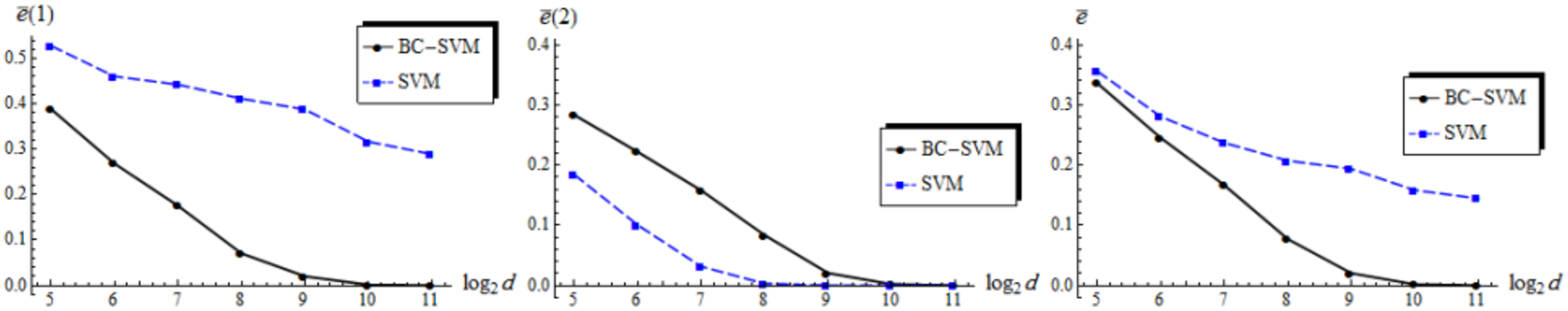} \\[-1mm]
(b) $(n_1,n_2)=(6,14)$ and $\bSig_1=\bSig_2=\bI_d$ (i.e., $\kappa/\Delta=6/7$) \\[2mm]
\includegraphics[scale=0.4]{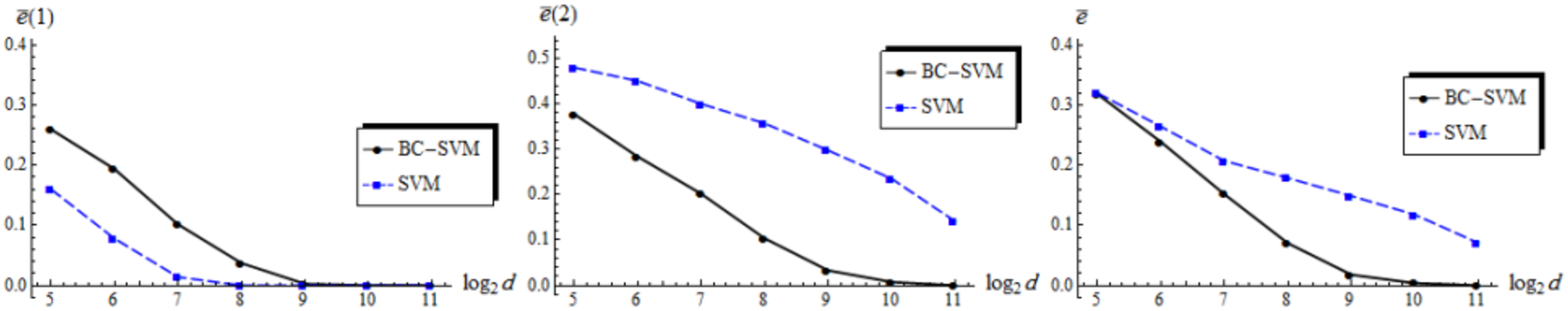} \\[-1mm]
(c) $(n_1,n_2)=(10,10)$, $\bSig_1=0.6\bI_d$ and $\bSig_2=1.4\bI_d$ (i.e., $\kappa/\Delta=-18/25$) 
\caption{The performance of the BC-SVM in HDLSS settings.
The error rates are denoted by the solid lines for (a), (b) and (c). 
The left panels display $\overline{e}(1)$, the middle panels display $\overline{e}(2)$ and the right panels display $\overline{e}$.
The corresponding error rates by the SVM are denoted by the dashed lines. 
\CG{Their standard deviations are less than $0.0112$.}
}
\label{F2}
\end{figure}

Next, we compared the performance of the BC-SVM with the SVM in complex settings. 
We set $\bmu_1=\bze$, 
$\bSig_1=\bB( 0.3^{|i-j|^{1/3}})\bB$ and $\bSig_2=\bB( 0.4^{|i-j|^{1/3}})\bB$, where
$$
\bB=\mbox{diag}[\{0.5+1/(d+1)\}^{1/2},...,\{0.5+d/(d+1)\}^{1/2}]. 
$$
Note that $\tr(\bSig_1)=\tr(\bSig_2)=d$. 
\CG{
We considered two cases: 
\begin{flushleft}
$\bmu_2=(1,...,1,0,...,0,-1,...,-1)^T$ $(=\bmu_{\alpha}(t),\ \mbox{say})$ whose first $t/2$ elements are $1$ and last $t/2$ elements are $-1$ for a positive even number $t$; and \\
$\bmu_2=(t^{1/2}/2,t^{1/2}/2,0,...,0,-t^{1/2}/2,-t^{1/2}/2)^T$ $(=\bmu_{\beta}(t),\ \mbox{say})$ whose 
first two elements are $t^{1/2}/2$ and last two elements are $-t^{1/2}/2$ for a positive number $t$. 
\end{flushleft}
Note that $\Delta=t$ both for $\bmu_{\alpha}(t)$ and $\bmu_{\beta}(t)$.}
We generated $\bx_{ij}-\bmu_i$, $i=1,2;\ j=1,2,...,$ independently either from (I) $N_d(\bze,\bSigma_i),\ i=1,2$, or (II) a $d$-variate $t$-distribution, $t_d(\bSig_i,10),\ i=1,2$, with mean zero, covariance matrix $\bSig_i$ and degrees of freedom 10. 
Note that (A-i) holds under (A-ii) for (I). 
Let $d_{*}=2\lceil d^{2/3}/2 \rceil$, where $ \lceil x \rceil$ denotes the smallest integer $\ge x$.
\CG{
We considered four cases: 
\begin{flushleft}
(d) $\bmu_2=\bmu_{\alpha}(d_{*})$, $(n_1,n_2)=(5,25)$ and $d=2^{s},\ s=6,...,12$, for (I);  \\
(e) $\bmu_2=\bmu_{\alpha}(d_{*})$, $d=1000$ and $(n_1,n_2)=(4s,8s),\ s=1,...,7$, for (II); \\ 
(f) $d=1000$, $(n_1,n_2)=(10,20)$ and $\bmu_2=\bmu_{\alpha}(2^s),\ s=1,...,7$, for (II); and \\ 
(g) $d=1000$, $(n_1,n_2)=(10,20)$ and $\bmu_2=\bmu_{\beta}(2^s),\ s=1,...,7$, for (II). \\ 
\end{flushleft}
Note that $\Delta=d_*=o(d)$ and (A-ii) holds for (d) and (e) from the fact that $\tr(\bSig_i^2)=O(d)$, $i=1,2$. 
Also, note that  (A-i) holds for (d). 
However, (A-i) does not hold for (e) and (A-iii) does not hold both for (d) and (e). 
For (f) and (g), we note that $\Delta=2^s,\ s=1,...,7$.}
\CG{ 
Especially, (g) is a sparse case such that the only four} \CG{elements of $\bmu_1-\bmu_2$ are nonzero.}
Similar to Section 1, we calculated the error rates, $\overline{e}(1)$, $\overline{e}(2)$ and $\overline{e}$, by 2000 replications and plotted the results in Figure 3.
\begin{figure}
\includegraphics[scale=0.4]{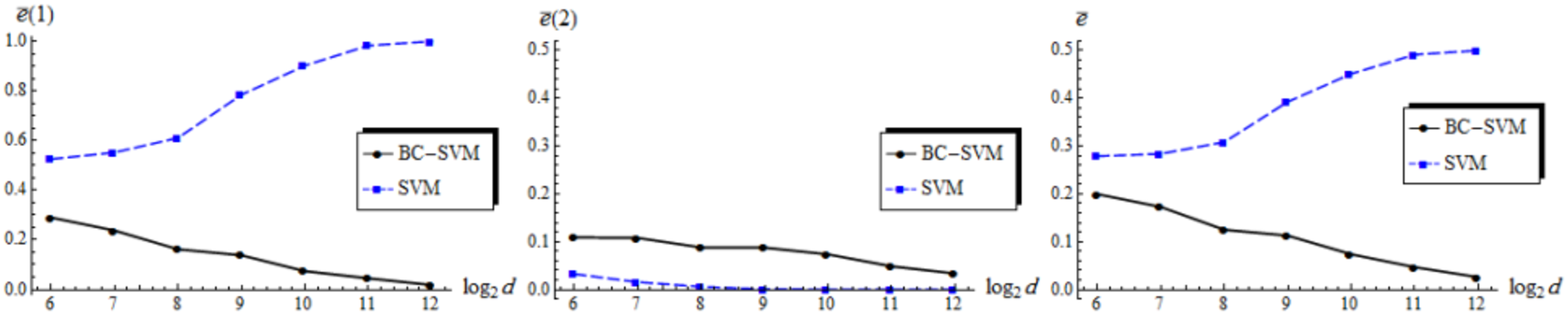} \\[-1mm]
{\footnotesize (d) $\bmu_2=\bmu_{\alpha}(d_{*})$ $(\Delta \approx d^{2/3})$, $(n_1,n_2)=(5,25)$ and $d=2^{s},\ s=6,...,12$, for (I) $N_d(\bze, \bSigma_i)$} \\[3mm]
\includegraphics[scale=0.4]{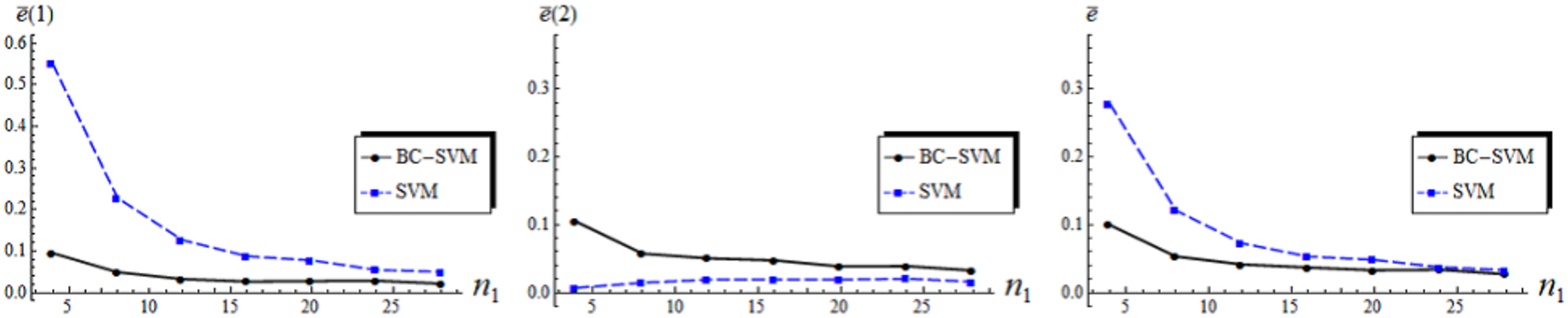} \\[-1mm]
{\footnotesize (e) $\bmu_2=\bmu_{\alpha}(d_{*})$ $(\Delta \approx d^{2/3})$, $d=1000$ and $(n_1,n_2)=(4s,8s),\ s=1,...,7$, for (II) $t_d(\bSig_i,10)$} \\[3mm]
\includegraphics[scale=0.403]{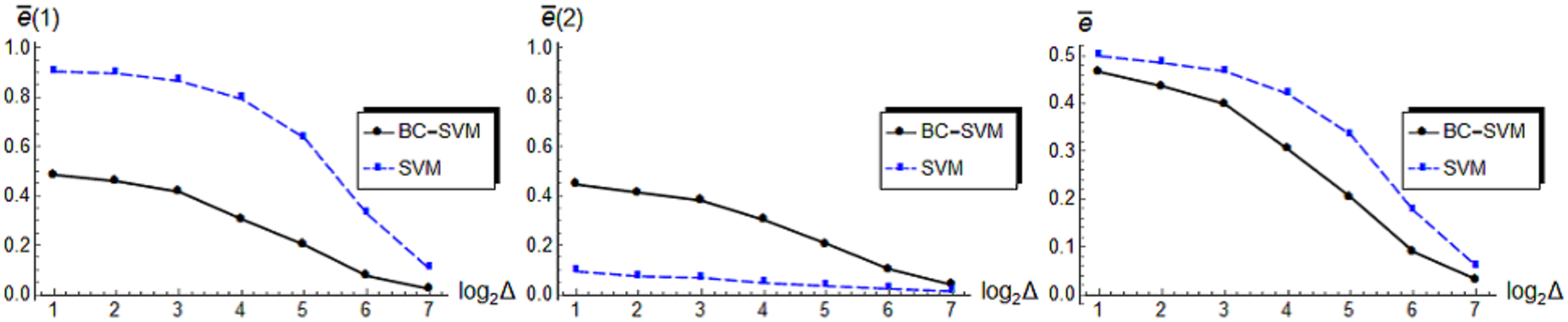} \\[-1mm]
{\footnotesize 
(f) $d=1000$, $(n_1,n_2)=(10,20)$ and 
$\bmu_2=\bmu_{\alpha}(2^s)\ (\Delta=2^s),\ s=1,...,7$, for (II) $t_d(\bSig_i,10)$ }\\[3mm] 
\includegraphics[scale=0.403]{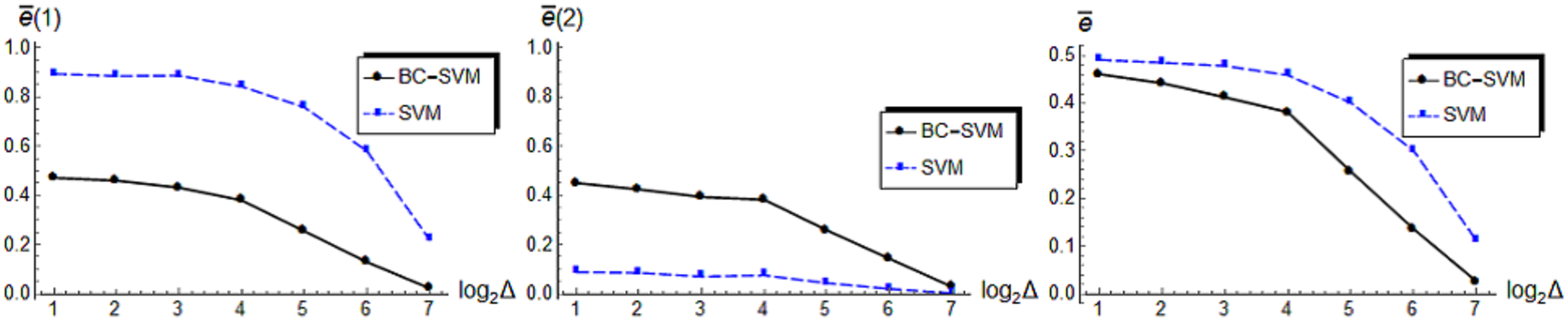} \\[-1mm]
{\footnotesize 
(g) $d=1000$, $(n_1,n_2)=(10,20)$ and 
$\bmu_2=\bmu_{\beta}(2^s)\ (\Delta=2^s),\ s=1,...,7$, for (II) $t_d(\bSig_i,10)$ }
\caption{The error rates of the BC-SVM and the SVM are denoted by the solid lines and the dashed lines, respectively, for (d) to (g).
The left panels display $\overline{e}(1)$, the middle panels display $\overline{e}(2)$ and the right panels display $\overline{e}$.
\CG{Their standard deviations are less than $0.0112$.}
}
\label{F3}
\end{figure}

We observe that the SVM gives quite bad performances for (d) in Figure 3. 
The main reason must be due to the bias term in the SVM. 
Note that $\kappa/\Delta \to \infty$ as $d\to \infty$ for (d). 
Thus $\overline{e}(1)$ becomes close to $1$ as $d$ increases. 
See Corollary 1 for the details. 
\CG{Also, the SVM gives bad performances for (e) to (g) when $n_i$s are small or $\Delta$ is small. 
This is because $\kappa/\Delta$ becomes large when $n_i$s are small or $\Delta$ is small. 
On the other hand, from Figures 2 and 3, the BC-SVM gives adequate performances 
even when $n_i$s and $\bSig_i$s are unbalanced.  
The BC-SVM also gives a better performance than the SVM even when $\Delta$ is small (or sparse).}
\subsection{Examples: Microarray data sets}
First, we used colon cancer data with $2000\ (=d)$ genes given by \cite{A99} which consists of $\pi_1:$ colon tumor (40 samples) and $\pi_2:$ normal colon (22 samples).
We set $n_1=n_2=10$. 
We randomly split the data sets from $(\pi_1,\pi_2)$ into training data sets of sizes $(n_1,n_2)$ and test data sets of sizes $(40-n_1,22-n_2)$. 
\CG{We constructed the BC-SVM and the SVM by using the training data sets. 
We checked accuracy by using the test data set for each $\pi_i$ and denoted the misclassification rates by $\widehat{e}(1)_r$ and $\widehat{e}(2)_r$. 
We repeated this procedure 100 times and obtained $\widehat{e}(1)_r$ and $\widehat{e}(2)_r$, $r=1,...,100$, both for the BC-SVM and the SVM.} 
We had the average misclassification rates as 
$\overline{e}(1)\ (=\sum_{r=1}^{100}\widehat{e}(1)_r/100)=0.16$, $\overline{e}(2)\ (=\sum_{r=1}^{100}\widehat{e}(2)_r/100)=0.166$ 
and $\overline{e}\ (=\{\overline{e}(1)+\overline{e}(2) \}/2)=0.163$ for the BC-SVM,
and $\overline{e}(1)=0.158$, $\overline{e}(2)=0.161$ and $\overline{e}=0.159$ for the SVM. 
By using all the samples, we considered estimating $\kappa/\Delta$. 
We set $m_1=40$ and $m_2=22$. 
From Section 3.1 in \cite{AY11}, an unbiased estimator of $\Delta$ was given by $\hat{\Delta}_{(m)}=\|\overline{\bx}_{1m_1}-\overline{\bx}_{2m_2} \|^2-\tr(\bS_{1m_1})/m_1-\tr(\bS_{2m_2})/m_2$. 
We estimated $\kappa/\Delta$ by 
$$
\widehat{\kappa/\Delta}=\{\tr(\bS_{1m_1})/n_1-\tr(\bS_{2m_2})/n_2\}/\hat{\Delta}_{(m)}
$$
and had $\widehat{\kappa/\Delta}=0.003$ for the $62$ samples.
In view of (\ref{3.1}), we expect that the BC-SVM is asymptotically equivalent to the SVM in such cases. 
\CG{We estimated $(\tr(\bSig_1)/\Delta,\tr(\bSig_2)/\Delta)$ by 
$(\tr(\bS_{1m_1})/\hat{\Delta}_{(m)},\tr(\bS_{2m_2})/\hat{\Delta}_{(m)})=(3.99,3.959)$.
It is difficult to estimate the standard deviation of the average misclassification rate. 
However, by noting that $\Var\{ \overline{e}(i)\}^{1/2}< \Var\{\widehat{e}(i)_r\}^{1/2}=[e(i)\{1-e(i)\}/(m_i-n_i)]^{1/2}$,   
one may have an upper bound of the standard deviation for $\overline{e}(i)$ as 
$$
s_u(i)=[\overline{e}(i)\{1-\overline{e}(i)\}/(m_i-n_i)]^{1/2},
$$
so that $\{ \sum_{i=1}^2s_u(i)^2/2 \}^{1/2}\ (=s_u,$ $\mbox{say})$ for $\overline{e}$. 
For the BC-SVM, $s_u(1)=0.067$, $s_u(2)=0.107$ and $s_u=0.089$.}
We summarized the results for various $n_i$s in Table 1. 
\begin{table}[tb]
 \caption{Average misclassification rates of the BC-SVM and the SVM, together with $\widehat{\kappa/\Delta}$, for \cite{A99}'s colon cancer data \CG{($d=2000$, $m_1=40$ and $m_2=22$). For each case, 
the standard deviations of $\overline{e}(1)$, $\overline{e}(2)$ and $\overline{e}$ 
are less than $s_u(1)$, $s_u(2)$ and $s_u$, respectively.}
}
 \begin{center}
  \begin{tabular}{c|ccc|ccc|c}
    \hline
       & \multicolumn{3}{c}{BC-SVM} & \multicolumn{3}{c}{SVM} & \\
 $(n_1,n_2)$ & $\overline{e}(1)$ &$\overline{e}(2)$ &$\overline{e}$& $\overline{e}(1)$ &$\overline{e}(2)$ &$\overline{e}$ 
 & $\widehat{\kappa/\Delta}$    \\
\hline
$(10,5)$  & $0.188$ & $0.209$ & $0.198$ & $0.122$ & $0.309$ & $0.215$ & $-0.393$ \\ 
$(10,10)$ & $0.16 $ & $0.166$ & $0.163$ & $0.158$ & $0.161$ & $0.159$ & $0.003$ \\
$(10,15)$ & $0.184$ & $0.156$ & $0.17$  & $0.206$ & $0.134$ & $0.17$  & $0.135$ \\
$(20,5)$  & $0.164$ & $0.249$ & $0.206$ & $0.082$ & $0.475$ & $0.278$ & $-0.592$ \\ 
$(20,10)$ & $0.141$ & $0.177$ & $0.159$ & $0.116$ & $0.23 $ & $0.173$ & $-0.196$ \\
$(20,15)$ & $0.142$ & $0.167$ & $0.154$ & $0.133$ & $0.181$ & $0.157$ & $-0.064$ \\ 
$(30,5)$  & $0.144$ & $0.302$ & $0.223$ & $0.083$ & $0.566$ & $0.324$ & $-0.659$ \\
$(30,10)$ & $0.12 $ & $0.236$ & $0.178$ & $0.108$ & $0.318$ & $0.213$ & $-0.263$ \\
$(30,15)$ & $0.115$ & $0.203$ & $0.159$ & $0.1  $ & $0.263$ & $0.181$ & $-0.131$ \\ 
       \hline
  \end{tabular}
 \end{center}
\end{table}

Next, we used leukemia data with $7129\ (=d)$ genes given by \cite{G99} which consists of $\pi_1:$ ALL (47\ \CG{$(=m_1)$} samples) and $\pi_2:$ AML (25 \CG{$(=m_2)$} samples). 
We applied the BC-SVM and the SVM to the leukemia data and summarized the results in Table 2. 
\CG{When $n_1 \neq n_2$, $|\widehat{\kappa/\Delta}|$ becomes large since
$(\tr(\bS_{1m_1})/\hat{\Delta}_{(m)},\tr(\bS_{2m_2})/\hat{\Delta}_{(m)})=(2.693,2.785)$.}
As expected theoretically, we observe that the BC-SVM gives adequate performances compared to the SVM when $|\widehat{\kappa/\Delta}|$ is not small. 
\begin{table}[tb]
 \caption{Average misclassification rates of the BC-SVM and the SVM, together with $\widehat{\kappa/\Delta}$, for \cite{G99}'s leukemia data ($d=7129$, $m_1=47$ and $m_2=25$). 
\CG{For each case, 
the standard deviations of $\overline{e}(1)$, $\overline{e}(2)$ and $\overline{e}$ 
are less than $s_u(1)$, $s_u(2)$ and $s_u$, respectively.}}
 \begin{center}
  \begin{tabular}{c|ccc|ccc|c}
    \hline
       & \multicolumn{3}{c}{BC-SVM} & \multicolumn{3}{c}{SVM}& \\
 $(n_1,n_2)$ & $\overline{e}(1)$ &$\overline{e}(2)$ &$\overline{e}$& $\overline{e}(1)$ &$\overline{e}(2)$ &$\overline{e}$ 
 & $\widehat{\kappa/\Delta}$    \\
\hline
$(10,5)$   & $0.044$ & $0.077$ & $0.06$  & $0.012$ & $0.148$ & $0.08$  & $-0.288$ \\
$(10,10)$  & $0.036$ & $0.043$ & $0.04$  & $0.036$ & $0.046$ & $0.041$ & $-0.009$ \\
$(10,20)$  & $0.044$ & $0.034$ & $0.039$ & $0.074$ & $0.026$ & $0.05$  & $0.13$   \\
$(20,5)$   & $0.031$ & $0.067$ & $0.049$ & $0.004$ & $0.199$ & $0.102$ & $-0.422$ \\
$(20,10)$  & $0.019$ & $0.051$ & $0.035$ & $0.011$ & $0.071$ & $0.041$ & $-0.144$ \\
$(20,20)$  & $0.028$ & $0.046$ & $0.037$ & $0.028$ & $0.046$ & $0.037$ & $-0.005$ \\ 
$(40,5)$   & $0.017$ & $0.102$ & $0.059$ & $0.0$   & $0.297$ & $0.149$ & $-0.49$  \\ 
$(40,10)$  & $0.016$ & $0.047$ & $0.031$ & $0.003$ & $0.091$ & $0.047$ & $-0.211$ \\  
$(40,20)$  & $0.011$ & $0.03$  & $0.021$ & $0.006$ & $0.032$ & $0.019$ & $-0.072$ \\
       \hline
  \end{tabular}
 \end{center}
\end{table}

\CG{Finally, we used myeloma data with $12625\ (=d)$ genes given by \cite{T03} which consists of $\pi_1:$ patients without bone lesions (36\ $(=m_1)$ samples) and $\pi_2:$ patients with bone lesions (137 $(=m_2)$ samples). 
We applied the BC-SVM and the SVM to the myeloma data and summarized the results in Table 3. 
When $n_1$ and $n_2$ are unbalanced, the SVM gives a very bad performance. 
This is because $\Delta$ in such cases is not sufficiently large since 
$(\tr(\bSig_1)/\Delta,\tr(\bSig_2)/\Delta)\approx (\tr(\bS_{1m_1})/\hat{\Delta}_{(m)},\tr(\bS_{2m_2})/\hat{\Delta}_{(m)})=(33.69,33.53)$, so that $\kappa/\Delta$ becomes too large when $n_1 \neq n_2$. 
Especially when $\widehat{\kappa/\Delta}>1$, $\overline{e}(1)$ of the SVM is too large. 
See Corollary 1 for the details. 
The BC-SVM also does not give a low error rate for this data because $\Delta$ is not sufficiently large. 
However, the BC-SVM gives adequate performances compared to the SVM especially when $\widehat{\kappa/\Delta}>1$.}
\CG{
Throughout Sections 3 and 4, we recommend to use the BC-SVM rather than the SVM for high-dimensional data.
}
\begin{table}[tb]
 \caption{\CG{Average misclassification rates of the BC-SVM and the SVM, together with $\widehat{\kappa/\Delta}$, for \cite{T03}'s myeloma data ($d=12625$, $m_1=36$ and $m_2=137$). 
For each case, 
the standard deviations of $\overline{e}(1)$, $\overline{e}(2)$ and $\overline{e}$ 
are less than $s_u(1)$, $s_u(2)$ and $s_u$, respectively.}
 }
 \begin{center}
  \begin{tabular}{c|ccc|ccc|c}
    \hline
       & \multicolumn{3}{c}{BC-SVM} & \multicolumn{3}{c}{SVM}& \\
 $(n_1,n_2)$ & $\overline{e}(1)$ &$\overline{e}(2)$ &$\overline{e}$& $\overline{e}(1)$ &$\overline{e}(2)$ &$\overline{e}$ 
 & $\widehat{\kappa/\Delta}$    \\
\hline
$(10,25)$  & $0.367$ & $0.307$ & $0.337$ & $0.787$  & $0.059$ & $0.423$ & $2.028$ \\ 
$(10,50)$  & $0.407$ & $0.265$ & $0.336$ & $0.936$  & $0.013$ & $0.475$ & $2.698$ \\ 
$(10,100)$ & $0.501$ & $0.193$ & $0.347$ & $0.993$  & $0.003$ & $0.498$ & $3.034$ \\ 
$(20,25)$  & $0.311$ & $0.288$ & $0.299$ & $0.401$  & $0.214$ & $0.308$ & $0.343$ \\
$(20,50)$  & $0.343$ & $0.25$  & $0.296$ & $0.646$  & $0.085$ & $0.365$ & $1.014$ \\
$(20,100)$ & $0.436$ & $0.175$ & $0.306$ & $0.872$  & $0.026$ & $0.449$ & $1.349$ \\
$(30,25)$  & $0.303$ & $0.288$ & $0.296$ & $0.25$   & $0.341$ & $0.295$ & $-0.218$\\
$(30,50)$  & $0.33$  & $0.26$ & $0.295$ & $0.467$  & $0.162$ & $0.314$ & $0.452$ \\
$(30,100)$ & $0.382$ & $0.195$ & $0.288$ & $0.713$  & $0.068$ & $0.391$ & $0.788$ \\
       \hline
  \end{tabular}
 \end{center}
\end{table}
\section{Multiclass SVMs}
In this section, we consider multiclass SVMs in HDLSS settings. 
We have i.i.d. observations, $\bx_{i1},...,\bx_{in_i}$, from each $\pi_i\ (i=1,...,g)$, where $g \ge 3$ and $\pi_i$ has a $d$-dimensional distribution with an unknown mean vector $\bmu_i$ and unknown covariance matrix $\bSigma_i\ (\ge \bO)$. 
We assume $n_i\ge 2,\ i=1,...,g$. 
Let $\Delta_{ij}=\|\bmu_i-\bmu_j\|^2$ for $i,j=1,...,g;\ i\neq j$.
We assume that $\tr(\bSig_i)/d \in (0,\infty)$ as $d\to \infty$ for $i=1,...,g$, 
and $\limsup_{d\to \infty}\Delta_{ij}/d<\infty $ for $i,j=1,...,g;\ i\neq j$. 
We consider the one-versus-one approach (the max-wins rule). 
See \cite{F96} and \cite{B06} for the details. 
\CG{Let $N_{g}=\sum_{i=1}^gn_i$. 
First, we consider the case when $d\to \infty$ while $N_{g}$ is fixed.} 
We consider the following assumptions: 
\begin{description}
  \item[(B-i)]\quad $\displaystyle \frac{ \max_{l=i,j}\Var(\|\bx_{lk}-\bmu_l\|^2) }{\Delta_{ij}^2}\to 0$ as $d\to \infty$ for $i,j=1,...,g;\ i\neq j$;    
  \item[(B-ii)]\quad $\displaystyle \frac{ \max_{l=i,j} \tr(\bSigma_l^2)}{\Delta_{ij}^2}\to 0$ as $d\to \infty$ for $i,j=1,...,g;\ i\neq j$.
\end{description}
Let $\kappa_{ij} =\tr(\bSig_i)/n_i-\tr(\bSig_j)/n_j$ for $i,j=1,...,g;\ i\neq j$. 
We consider the following condition: 
\begin{description}
  \item[(B-iii)]\quad $\displaystyle  \limsup_{d\to \infty} \frac{|\kappa_{ij}|}{\Delta_{ij}}<1$ 
  for $i,j=1,...,g;\ i\neq j$.
\end{description}
From Theorem 1, for the one-versus-one approach by (\ref{2.4}), we have the following result. 
\begin{cor}
Under (B-i) to (B-iii), it holds for the multiclass SVM that 
\begin{equation}
e(i)\to 0 \ \ \mbox{as $d\to \infty$ for $i=1,...,g$}.
\label{5.1}
\end{equation}
\end{cor}
From Theorem 2, for the one-versus-one approach by (\ref{3.1}), we have the following result. 
\begin{cor}
Under (B-i) and (B-ii), the multiclass BC-SVM holds (\ref{5.1}). 
\end{cor}
Note that the BC-SVM satisfies the consistency property without (B-iii). 
Thus we recommend to use the BC-SVM in multiclass HDLSS settings. 

\CG{Next, we consider the case when both $d,N_{g}\to \infty$ while $N_{g}/d\to 0$. 
Similar to Section 2.3 and Corollary 3, 
the multiclass SVMs have the consistency property 
under some regularity conditions.} 

\CG{
We checked the performance of the multiclass SVMs by using leukemia data with $12582\ (=d)$ genes given by 
\cite{A02} which consists of $\pi_1:$ ALL (24\ $(=m_1)$ samples), 
$\pi_2:$ MLL (20\ $(=m_2)$ samples) and 
$\pi_3:$ AML (28\ $(=m_3)$ samples). 
We applied the multiclass BC-SVM and SVM to the leukemia and summarized the results in Table 4.
We had $(\tr(\bS_{1m_1})/\hat{\Delta}_{12(m)},$ $\tr(\bS_{2m_2})/\hat{\Delta}_{12(m)})=(2.724, 3.213)$, 
$(\tr(\bS_{1m_1})/\hat{\Delta}_{13(m)},\tr(\bS_{3m_3})/\hat{\Delta}_{13(m)})=(0.738,0.9)$ 
and $(\tr(\bS_{2m_2})/\hat{\Delta}_{23(m)},\tr(\bS_{3m_3})/\hat{\Delta}_{23(m)})=(1.533,1.585)$, 
where $\hat{\Delta}_{ij(m)}=\|\overline{\bx}_{im_i}-\overline{\bx}_{jm_j} \|^2-\tr(\bS_{im_i})/m_i-\tr(\bS_{jm_j})/m_j$ that is an unbiased estimator of $\Delta_{ij}$. 
Thus $|\kappa_{ij}/\Delta_{ij}|$ must become large when $n_i\neq n_j$. 
Actually, the multiclass BC-SVM gives adequate performances for all the cases.}
\begin{table}[tb]
\caption{\CG{
Average misclassification rates of the BC-SVM and the SVM for \cite{A02}'s leukemia data ($d=12582$, $m_1=24$, $m_2=20$ and $m_3=28$). 
For each case, 
the standard deviations of $\overline{e}(i)$, $i=1,2,3,$ and $\overline{e}$ 
are less than $s_u(i)$, $i=1,2,3,$ and $s_u=\{ \sum_{i=1}^3s_u(i)^2/3 \}^{1/2}$, respectively.}}
 \begin{center}
   \begin{tabular}{c|cccc|cccc}
    \hline
       & \multicolumn{4}{c}{BC-SVM} & \multicolumn{4}{c}{SVM} \\
 $(n_1,n_2,n_3)$ & $\overline{e}(1)$ &$\overline{e}(2)$ & $\overline{e}(3)$ & $\overline{e}$& $\overline{e}(1)$ &$\overline{e}(2)$ & $\overline{e}(3)$ &$\overline{e}$    \\
\hline
$(5,5,10)$  & $0.085$  & $0.089$ & $0.071$ & $0.082$ & $0.069$  & $0.118$ & $0.06$ & $0.082$ \\
$(5,5,20)$  & $0.103$  & $0.087$ & $0.07$ & $0.087$ & $0.089$  & $0.135$ & $0.053$ & $0.092$ \\
$(5,10,10)$ & $0.049$  & $0.06$ & $0.066$ & $0.058$ & $0.095$  & $0.047$ & $0.066$ & $0.069$ \\
$(5,10,20)$ & $0.044$  & $0.068$ & $0.064$ & $0.059$ & $0.088$  & $0.06$ & $0.06$ & $0.069$ \\
$(10,5,10)$ & $0.051$  & $0.077$ & $0.063$ & $0.064$ & $0.021$  & $0.143$ & $0.049$ & $0.071$ \\
$(10,5,20)$ & $0.051$  & $0.073$ & $0.061$ & $0.062$ & $0.018$  & $0.148$ & $0.044$ & $0.07$ \\
$(10,10,10)$ & $0.028$  & $0.056$ & $0.063$ & $0.049$ & $0.025$  & $0.059$ & $0.064$ & $0.049$ \\
$(10,10,20)$ & $0.031$  & $0.051$ & $0.071$ & $0.051$ & $0.03$  & $0.058$ & $0.065$ & $0.051$ \\
       \hline
  \end{tabular}
 \end{center}
\end{table}
\appendix
\section{}
Throughout, let $\bmu=\bmu_1-\bmu_2$ and $\bmu_*=(\bmu_1+\bmu_2)/2$.
\begin{proof}[Proof of Lemma 1] 
Under (A-ii), we have that as $d\to \infty$
\begin{align}
\bmu^T\bSig_i\bmu/\Delta^2 \le \tr(\bSig_i^2)^{1/2}/\Delta=o(1),\quad i=1,2.
\label{A.1}
\end{align}
Then, by using Chebyshev's inequality, for any $\tau>0$, under (A-ii), we have that 
\begin{align}
&P(|(\bx_j-\bmu_*)^T(\bx_k-\bmu_*)-\Delta/4 |\ge \tau \Delta )\notag \\
& \le (\tau \Delta)^{-2}E[\{(\bx_j-\bmu_*)^T(\bx_k-\bmu_*)-\Delta/4 \}^2] \notag \\
&=O\{\tr(\bSig_1^2)+\bmu^T\bSig_1\bmu\}/\Delta^2=o(1)\quad \mbox{for $1\le j< k\le n_1$}; \notag \\
&P(|(\bx_j-\bmu_*)^T(\bx_k-\bmu_*)-\Delta/4 |\ge \tau \Delta ) \notag \\
&=O\{\tr(\bSig_2^2)+\bmu^T\bSig_2\bmu\}/\Delta^2=o(1)\quad 
\mbox{for $n_1+1 \le j< k\le N$};\quad \mbox{and} \notag \\
&P(|(\bx_j-\bmu_*)^T(\bx_k-\bmu_*)+\Delta/4 |\ge \tau \Delta )\notag \\
&=O\{\tr(\bSig_1\bSig_2)+\bmu^T(\bSig_1+\bSig_2)\bmu\}/\Delta^2=o(1)\notag \\
&\quad \ \mbox{for $1 \le j \le n_1$ and $n_1+1\le k\le N$} 
\label{A.2}
\end{align}
from the fact that $\tr(\bSig_1\bSig_2)\le \{\tr(\bSig_1^2)\tr(\bSig_2^2)\}^{1/2}$.
From (\ref{A.1}), for any $\tau>0$, we have that 
\begin{align}
&P(| \|\bx_j-\bmu_*\|^2-\Delta/4-\tr(\bSig_1) |\ge \tau \Delta )\notag\\
&=O\{\mbox{Var}(\|\bx_{1j}-\bmu_1\|^2) +\bmu^T\bSig_1\bmu\}/\Delta^2=o(1) \quad \mbox{for $j=1,...,n_1$}; \quad and\notag \\
&P(|\|\bx_j-\bmu_*\|^2-\Delta/4-\tr(\bSig_2) |\ge \tau \Delta )=o(1) \quad \mbox{for $j=n_1+1,...,N$}
\label{A.3}
\end{align}
under (A-i) and (A-ii).
Here, subject to (\ref{2.3}), we can write for (\ref{2.2}) that
\begin{equation}
L(\balpha)=\sum_{j=1}^N\alpha_j-\frac{1}{2}\sum_{j=1}^N\sum_{k=1}^N\alpha_j\alpha_kt_jt_k (\bx_j-\bmu_*)^T(\bx_k-\bmu_*).
\label{A.4}
\end{equation}
Then, by noting that $\alpha_j\ge 0$ for all $j$ subject to (\ref{2.3}), from (\ref{A.2}) and (\ref{A.3}), we have that 
\begin{align}
L(\balpha)=&\sum_{j=1}^N \alpha_j- \frac{\Delta}{8}\Big(\sum_{j=1}^N \alpha_j\Big)^2-
\frac{1}{2}\Big(\tr(\bSig_1)\sum_{j=1}^{n_1}\alpha_j^2+
\tr(\bSig_2)\sum_{j=n_1+1}^{N}\alpha_j^2\Big)\notag \\
&+o_p\Big\{\Delta\Big(\sum_{j=1}^N \alpha_j\Big)^2  \Big\}\label{A.5}
\end{align}
subject to (\ref{2.3}) under (A-i) and (A-ii).
It concludes the result.
\end{proof}
\begin{proof}[Proof of Lemma 2] 
By combining Lemma 1 with (\ref{2.5}) and (\ref{2.6}), we can claim the first result.

When $\hat{S}=\{1,...,N\}$, by noting that $\sum_{j=1}^N\hat{\alpha}_jt_j=0$, we have that 
\begin{align}
\hat{y}(\bx_0)=&\sum_{j=1}^N \hat{\alpha}_jt_j (\bx_j-\bmu_*)^T(\bx_0-\bmu_*)+\sum_{j=1}^N \hat{\alpha}_jt_j (\bx_j-\bmu_*)^T\bmu_*+\hat{b} \notag \\
=&\sum_{j=1}^N \hat{\alpha}_jt_j (\bx_j-\bmu_*)^T(\bx_0-\bmu_*)\notag \\ 
&+
\frac{-n_1+n_2}{N}-\frac{1}{N}\sum_{j=1}^N\sum_{k=1}^N \hat{\alpha}_kt_k(\bx_j-\bmu_*)^T(\bx_k-\bmu_*).  \label{A.6}
\end{align}
From the first result of Lemma 2, (\ref{A.2}) and (\ref{A.3}), we have that as $d\to \infty$
\begin{align}
&\frac{-n_1+n_2}{N}-\frac{1}{N}\sum_{j=1}^N\sum_{k=1}^N \hat{\alpha}_kt_k(\bx_j-\bmu_*)^T(\bx_k-\bmu_*)\notag \\
&=\frac{-n_1+n_2}{N}+\frac{(n_1-n_2)\Delta}{\Delta_*N}+2
\frac{\tr(\bSig_1)-\tr(\bSig_2)}{\Delta_*N}+o_p\Big( \frac{\Delta}{\Delta_*} \Big) \notag \\
&=\frac{-n_1+n_2}{N}\Big(\frac{\delta}{\Delta_*} \Big)+2
\frac{\tr(\bSig_1)-\tr(\bSig_2)}{\Delta_*N}+o_p\Big( \frac{\Delta}{\Delta_*} \Big)  \notag \\
&=
\frac{\tr(\bSig_1)/n_1-\tr(\bSig_2)/n_2}{\Delta_*}+o_p
\Big( \frac{\Delta}{\Delta_*} \Big) 
\label{A.7} 
\end{align}
under (A-i) and (A-ii). 
Similar to (\ref{A.2}), under (A-ii), we obtain that $(\bx_{j}-\bmu_*)^T(\bx_0-\bmu_*)/\Delta=(-1)^{i+1}/4+o_p(1)$ for $j=1,...,n_1$, and $(\bx_{j}-\bmu_*)^T(\bx_0-\bmu_*)/\Delta=(-1)^{i}/4+o_p(1)$ for $j=n_1+1,...,N$, when $\bx_0 \in \pi_i$ $(i=1,2)$. 
Then, from the first result of Lemma 2, under (A-i) and (A-ii), 
it holds that 
\begin{align}
\sum_{j=1}^N\hat{\alpha}_jt_j (\bx_j-\bmu_*)^T(\bx_0-\bmu_*)=\frac{(-1)^i\Delta}{\Delta_*}
+o_p\Big( \frac{\Delta}{\Delta_*} \Big) \label{A.8} 
\end{align}
when $\bx_0 \in \pi_i$ for $i=1,2$. 
By combining (\ref{A.6}) with (\ref{A.7}) and (\ref{A.8}), 
we can conclude the second result.
\end{proof}
\begin{proof}[Proofs of Theorem 1 and Corollary 1]
By using (\ref{2.7}), the results are obtained straightforwardly. 
\end{proof}
\CG{
\begin{proof}[Proof of Lemma 3]
Similar to (\ref{A.2}), under (A-ii'),
from (\ref{A.1}), we have that as $d,N\to \infty$
\begin{align}
&\sum_{1\le j< k\le n_1} P(|(\bx_j-\bmu_1)^T(\bx_k-\bmu_1)|\ge \tau \Delta )=O\Big(\frac{n_1^2\tr(\bSig_1^2)}{\Delta^2} \Big)=o(1);\notag \\
&\sum_{n_1+1 \le j< k\le N} P(|(\bx_j-\bmu_2)^T(\bx_k-\bmu_2) |\ge \tau \Delta )=O\Big(\frac{n_2^2\tr(\bSig_2^2)}{\Delta^2} \Big)=o(1);\notag \\
&\sum_{j=1}^{n_1}\sum_{k=n_1+1}^{N} P(|(\bx_j-\bmu_1)^T(\bx_k-\bmu_2) |\ge \tau \Delta )=O\Big(\frac{n_1n_2\tr(\bSig_1\bSig_2)}{\Delta^2} \Big)=o(1);\notag \\
&\sum_{j=1}^{n_1} P(|(\bx_j-\bmu_1)^T\bmu |\ge \tau \Delta )=O\Big(\frac{n_1\bmu^T\bSig_1\bmu  }{\Delta^2} \Big)=
O\Big(\frac{n_1\tr(\bSig_1^2)^{1/2}}{\Delta} \Big)=o(1); \notag \\
&\mbox{and}\quad 
\sum_{j=n_1+1}^{N} P(|(\bx_j-\bmu_2)^T\bmu |\ge \tau \Delta )=O\Big(\frac{n_2\tr(\bSig_2^2)^{1/2}}{\Delta} \Big) =o(1) \notag 
\end{align}
for any $\tau>0$. 
Then, under (A-ii'), we have that 
\begin{align}
&(\bx_j-\bmu_*)^T(\bx_k-\bmu_*)=\Delta\{1+o_p(1)\}/4\quad \mbox{for all $1\le j< k\le n_1$}; \notag \\
&(\bx_j-\bmu_*)^T(\bx_k-\bmu_*)=\Delta\{1+o_p(1)\}/4\quad \mbox{for all $n_1+1\le j< k\le N$};\quad \mbox{and}  \notag \\
&(\bx_j-\bmu_*)^T(\bx_k-\bmu_*)=-\Delta\{1+o_p(1)\}/4\notag \\
& \mbox{for all  $1 \le j \le n_1$ and $n_1+1\le k\le N$}. \label{A.9}
\end{align}
On the other hand, for any $\tau>0$, we have that 
$\sum_{j=1}^{n_1} P(| \|\bx_j-\bmu_*\|^2-\Delta/4-\tr(\bSig_1) |\ge \tau \Delta )= O\{n_1 \mbox{Var}(\|\bx_{1j}-\bmu_1\|^2) +n_1 \bmu^T\bSig_1\bmu\}/\Delta^2=o(1)$ and 
$\sum_{j=n_1+1}^{N}$
$ P(|\|\bx_j-\bmu_*\|^2-\Delta/4-\tr(\bSig_2) |\ge \tau \Delta )=o(1)$ 
under (A-i') and (A-ii') as $d,N\to \infty$, 
so that 
\begin{align}
& \|\bx_j-\bmu_*\|^2=\Delta\{1+o_p(1)\}/4+\tr(\bSig_1) \quad \mbox{for all $1\le j \le n_1$};\quad \mbox{and} \notag \\
& \|\bx_j-\bmu_*\|^2=\Delta\{1+o_p(1)\}/4+\tr(\bSig_2) \quad \mbox{for all $n_1+1 \le j \le N$}. \label{A.10}
\end{align}
Then, by combining (\ref{A.4}) with (\ref{A.9}) and (\ref{A.10}), we have (\ref{A.5}) as $d,N\to \infty$, 
subject to (\ref{2.3}) under (A-i') and (A-ii'). 
Similar to the proof of Lemma 2, 
by noting (A-iv), we can conclude the result. 
\end{proof}
}
\begin{proof}[Proof of Lemma 4]
We have that 
\begin{align}
\hat{\Delta}_*-{\Delta}_*=&
\sum_{i=1}^2\sum_{j=1}^{n_i}\frac{\|\bx_{ij}-\bmu_i\|^2-\tr(\bSig_i) }{n_i^2}+
\sum_{i=1}^2\sum_{j\neq k}^{n_i}\frac{(\bx_{ij}-\bmu_i)^T(\bx_{ik}-\bmu_i)}{n_i^2} \notag \\
&+\sum_{i=1}^2(-1)^{i+1}\bmu^T(\overline{\bx}_{in_i}-\bmu_i)-2(\overline{\bx}_{1n_1}-\bmu_1)^T
(\overline{\bx}_{2n_2}-\bmu_2). \label{A.11}
\end{align}
Note that $E[\{\|\bx_{ij}-\bmu_i\|^2-\tr(\bSig_i)\}^2]=o(\Delta^2)$ as $d\to \infty$ under (A-i) for all $i,j$. 
Also, 
note that $E[\{\bmu^T(\overline{\bx}_{in_i}-\bmu_i)\}^2]=\bmu^T\bSig_i\bmu/n_i\le \Delta \tr(\bSig_i^2)^{1/2}/ n_i=o(\Delta^2/n_i)$ as $d\to \infty$ under (A-ii) for $i=1,2$. 
Then, from (\ref{A.11}), we can claim that $E\{(\hat{\Delta}_*-{\Delta}_*)^2\}=o(\Delta^2)$ under (A-i) and (A-ii), 
so that $\hat{\Delta}_*={\Delta}_*+o_p(\Delta)$. 
On the other hand, we have that 
$$
\tr(\bS_{in_i})-\tr(\bSig_i)=\sum_{j=1}^{n_i}\frac{\|\bx_{ij}-\bmu_i\|^2-\tr(\bSig_i) }{n_i}
-\sum_{j\neq k}^{n_i}\frac{(\bx_{ij}-\bmu_i)^T(\bx_{ik}-\bmu_i)}{n_i(n_i-1)}.
$$
Then, similar to $\hat{\Delta}_*$, we can claim that $\tr(\bS_{in_i})=\tr(\bSig_i)+o_p(\Delta)$ for $i=1,2$, under (A-i) and (A-ii), so that $\hat{\kappa}=\kappa+o_p(\Delta)$. 
Hence, by noting that $|\kappa|/\Delta_*\le 1$, we can claim the result. 
\end{proof}
\begin{proof}[Proof of Theorem 2]
By using (\ref{3.2}), the result is obtained straightforwardly. 
\end{proof}
\CG{
\begin{proof}[Proofs of Corollaries 2 and 3]
From Lemma 3, we have (\ref{2.7}) as $d,N\to \infty$ under (A-i'), (A-ii') and (A-iv). 
We note that Lemma 4 holds even when $d,N\to \infty$. 
Hence, from (\ref{2.7}) and Lemma 4, we can claim the results.
\end{proof}
}
\begin{proof}[Proofs of Corollaries 4 and 5]
By using Theorems 1 and 2, the results are obtained straightforwardly. 
\end{proof}
\section*{Acknowledgements}
Research of the second author was partially supported by 
Grant-in-Aid for Young Scientists (B), Japan Society for the Promotion of Science (JSPS), under Contract Number 26800078.
Research of the third author was partially supported by 
Grants-in-Aid for Scientific Research (A) and Challenging Exploratory Research, JSPS, under Contract Numbers 15H01678 and 26540010.

\end{document}